\documentclass{article}

\usepackage{amsmath,amsfonts,bm}

\def\eqref#1{equation~\ref{#1}}

\def\1{\bm{1}}

\def\rvx{{\mathbf{x}}}

\def\rvz{{\mathbf{z}}}

\DeclareMathAlphabet{\mathsfit}{\encodingdefault}{\sfdefault}{m}{sl}
\SetMathAlphabet{\mathsfit}{bold}{\encodingdefault}{\sfdefault}{bx}{n}

\newcommand{\E}{\mathbb{E}}

\newcommand{\KL}{D_{\mathrm{KL}}}

\usepackage{amssymb}
\usepackage{amsmath}
\usepackage{graphicx}
\usepackage{nicefrac}
\usepackage{algorithm}
\usepackage{algorithmic}
\usepackage{enumitem}

\usepackage[final]{corl_2025} 
\usepackage{multirow}
\usepackage{mathabx}

\usepackage[dvipsnames]{xcolor}
\definecolor{customlinkcolor}{HTML}{2774AE} 
\definecolor{customcitecolor}{HTML}{2774AE} 

\hypersetup{
    colorlinks=true,
    linkcolor=customlinkcolor, 
    anchorcolor=black, 
    citecolor=customcitecolor  
}
\usepackage[capitalize,noabbrev]{cleveref}

\title{Latent Adaptive Planner for Dynamic Manipulation}

\author{
  Donghun Noh$^{1\spadesuit}$, Deqian Kong$^{1,2\spadesuit}$, Minglu Zhao$^{1}$, Andrew Lizarraga$^{1}$, \\\textbf{Jianwen Xie$^{2}$, Ying Nian Wu$^{1\clubsuit}$, Dennis Hong$^{1\clubsuit}$}\\
  $^1$UCLA\quad$^2$Lambda, Inc.\\ 
  $^\spadesuit$Equal contribution\quad$^\clubsuit$Equal advising\\ 
}

\begin{document}
\maketitle


\begin{abstract}
    We present the Latent Adaptive Planner (LAP), a trajectory‑level latent‑variable policy for dynamic nonprehensile manipulation (e.g., box catching) that formulates planning as inference in a low‑dimensional latent space and is learned effectively from human demonstration videos. During execution, LAP achieves real-time adaptation by maintaining a posterior over the latent plan and performing variational replanning as new observations arrive.
    To bridge the embodiment gap between humans and robots, we introduce a model-based proportional mapping that regenerates accurate kinematic-dynamic joint states and object positions from human demonstrations. 
    Through challenging box catching experiments with varying object properties, LAP demonstrates superior success rates, trajectory smoothness, and energy efficiency by learning human-like compliant motions and adaptive behaviors.
    Overall, LAP enables dynamic manipulation with real-time adaptation and successfully transfer accross heterogeneous robot platforms using the same human demonstration videos.
\end{abstract}



\section{Introduction}
Dynamic manipulation, which involves controlling objects through rapid contact changes and complex physical interactions~\citep{mason1993dynamic, ruggiero2018nonprehensile}, remains a fundamental challenge in robotics. While humans naturally perform throwing, catching, and rapid transfers, robots remain confined to slow, conservative movements. Achieving true dynamic manipulation requires real-time consideration of diverse physical properties of objects (mass, friction, elasticity)~\citep{noh2021surface, noh2024control}, changes in contact states, and joint torques~\citep{lynch1999dynamic, tsuji2015dynamic, gong2023legged}, yet these dynamic characteristics are difficult to model analytically and vary significantly across different objects~\citep{yu2016more, fazeli2017parameter}. Recent advances like Tossingbot~\citep{zeng2020tossingbot}, FlingBot~\citep{ha2022flingbot}, and Chi et al.~\citep{chi2024iterative} have shown progress in specific scenarios, yet predominantly rely on self-supervised learning requiring thousands of robot trials. For this reason, this approach lacks scalability since exploration with heavier objects or mobile-based platforms like humanoids become unsafe and leads to significant hardware damage upon failure. While simulation offers safer training, accurately modeling dynamic manipulation physics remains challenging, limiting sim-to-real transfer.

Recently, imitation learning has shown promising results on static or quasi-static manipulation tasks. However, recent approaches such as diffusion policy are limited by slow inference speeds that cannot meet the real-time requirements of dynamic manipulation. Moreover, most datasets for policy training consider only end-effector positions while overlooking critical dynamic elements such as contact information, joint torques, and grasping forces~\citep{chi2023diffusion, lepert2025phantom, shridhar2022cliport}. While more accurate data can be collected through teleoperation, this approach presents practical barriers including high implementation costs, logistical challenges for achieving scale and diversity, and technical difficulties in capturing complete physical interaction data~\citep{mandlekar2020iris, zhang2018deep, mandlekar2019scaling}. Additionally, demonstration-based learning approaches struggle with temporal consistency and trajectory smoothness when transferring skills from human demonstrations to robotic systems~\citep{rajeswaran2017learning}. Although recent generative modeling approaches such as sequence modeling~\citep{chen2021decision} and diffusion models~\citep{chi2023diffusion} have addressed some of these challenges, they still exhibit limitations in handling the long-term dependencies and environmental contingencies inherent in dynamic manipulation tasks~\citep{mandlekar2019scaling}.

In this work, we introduce Latent Adaptive Planner (LAP), which formulates planning as latent space inference. Our approach addresses key challenges in visuomotor policy learning through a principled variational replanning framework that maintains temporal consistency while efficiently adapting to environmental changes. By leveraging a latent variable that functions as an abstract plan, LAP enables more coherent long-horizon planning while maintaining the flexibility needed for real-time adaptation. The system employs Bayesian updating in latent space to incrementally refine plans as new observations become available, balancing computational efficiency with real-time adaptability. Furthermore, by model-based proportional mapping to regenerate accurate kinematic-dynamic joint states and object positions from human demonstration videos, our approach bridges the gap between human motion and robotic execution, capturing the adaptive and energy-efficient qualities inherent in human manipulation skills. Importantly, this regeneration methodology allows us to create versatile datasets suitable for training diverse robotic platforms on the same manipulation tasks without requiring direct robot data collection.

We demonstrate our approach through box catching, one of the most challenging reactive manipulation tasks. Unlike controlled release scenarios such as throwing and fling, catching requires real-time adaptation to unpredictable trajectories while managing objects without secure grasps~\citep{kim2014catching, lampariello2011trajectory}. Boxes which have non-trivial weight and volumes tumble chaotically with asymmetric drag, forcing robots to predict contact locations and orientations while coordinating dual-arm movements with millisecond-scale precision and determining appropriate contact forces to absorb impact without dropping the object~\citep{kim2014catching, yan2024impact}. Through this task, we show that imitation learning can achieve energy-efficient dynamic manipulation while bypassing the intractable formulations required by optimization methods, demonstrating the potential for broader dynamic manipulation challenges.


\section{Related Works}
\label{sec:relatedworks}
\paragraph{Human Demonstration-based Robot Data Generation}
Those approaches that utilize human videos offer greater scalability but face embodiment gaps between humans and robots. Some methods address this by requiring hybrid datasets with both human and robot demonstrations~\cite{ebert2021bridge, sharma2018multiple}. Others extract action labels from videos through object tracking~\cite{song2020grasping, mandikal2022dexvip, cai2016understanding, chen2021joint} or inverse modeling techniques~\cite{young2021visual, schmeckpeper2020learning, schmeckpeper2020reinforcement}. Cross-embodiment techniques like RoviAug~\cite{chen2024rovi} but when applied to human-to-robot transfer~\cite{EgoMimic}, typically still require robot data.

Recent work has explored more efficient ways to leverage human demonstrations. Phantom~\cite{Phantom} enables zero-shot transfer of policies trained solely on human demonstrations to robot embodiments through simple data editing techniques, eliminating the need for robot data collection. Similarly, DROID~\cite{DROID} demonstrates that collecting diverse manipulation data across multiple environments creates more robust policies capable of generalizing to new scenarios. DexMV~\cite{DexMV} introduced a particularly relevant approach by establishing a framework for learning dexterous manipulation from human videos through demonstration translation techniques. This approach extracts 3D hand and object poses from videos and maps human hand trajectories to robot joint torques via inverse dynamics, closely aligning with our method of processing video data to generate robot demonstrations. Our specific data generation process, which involves human pose estimation, object pose extraction, and robot-specific scaling and kinematic retargeting, is detailed in~\cref{subsec:data}.

\paragraph{Imitation Learning and Latent Space Planning} Imitation learning enables robots to acquire skills by mimicking demonstrations, transforming complex control into supervised learning problems that map observations to actions~\citep{schaal1999imitation}, though traditional approaches like Behavior Cloning struggle with distribution shift during deployment~\citep{argall2009survey}. Recent approaches have reframed decision-making as sequence modeling~\cite{chen2021decision,janner2021offline} to capture temporal dependencies in state-action sequences.
Diffusion models offer an alternative paradigm for imitation learning, with~\citet{chi2023diffusion} generating multi-modal action distributions through iterative denoising processes and~\citet{janner2022planning} leveraging classifier-guided diffusion models.

Latent variable models address temporal consistency challenges by encoding trajectory-level information that captures long-term dependencies. \citet{yang2022dichotomy,paster2022you} investigate Decision Transformer's overfitting to environment contingencies and propose latent variable solutions that encode trajectory-level information.  Latent Plan Transformer~\citep{kong2024latent} formulates planning as latent space inference and decouples trajectory generation from return estimation, enabling more coherent long-horizon planning. Earlier works~\citep{ajay2021opal,lynch2020learning} propose VAE-based models~\cite{kingma2013auto} for temporally extended policies, but lack the adaptive replanning capabilities central to our framework. Our work builds upon these foundations but introduces classical variational Bayes learning~\citep{jordan1999introduction,blei2017variational} for precise plan inference and principled Bayesian updating in latent space, enabling efficient plan refinement as new observations become available as detailed in~\cref{sec:lap}.

\section{Method}
\label{sec:method}

The Latent Adaptive Planner (LAP) presents a novel methodology for dynamic manipulation tasks, addressing the inherent challenges in human demonstration learning and real-time environmental adaptation. Within this framework, effective data regeneration from human demonstrations is achieved, and adaptive responses to dynamic environments in real-time are enabled. The integration of latent space representation with variational inference allows for precise mapping between demonstration data and robot execution parameters while maintaining adaptability to unpredicted environmental changes.

\subsection{Robot Model-Based Data Regeneration from Human Demonstration Videos}
\label{subsec:data}
\begin{figure}[t!]
    \centering
    \includegraphics[width=\linewidth]{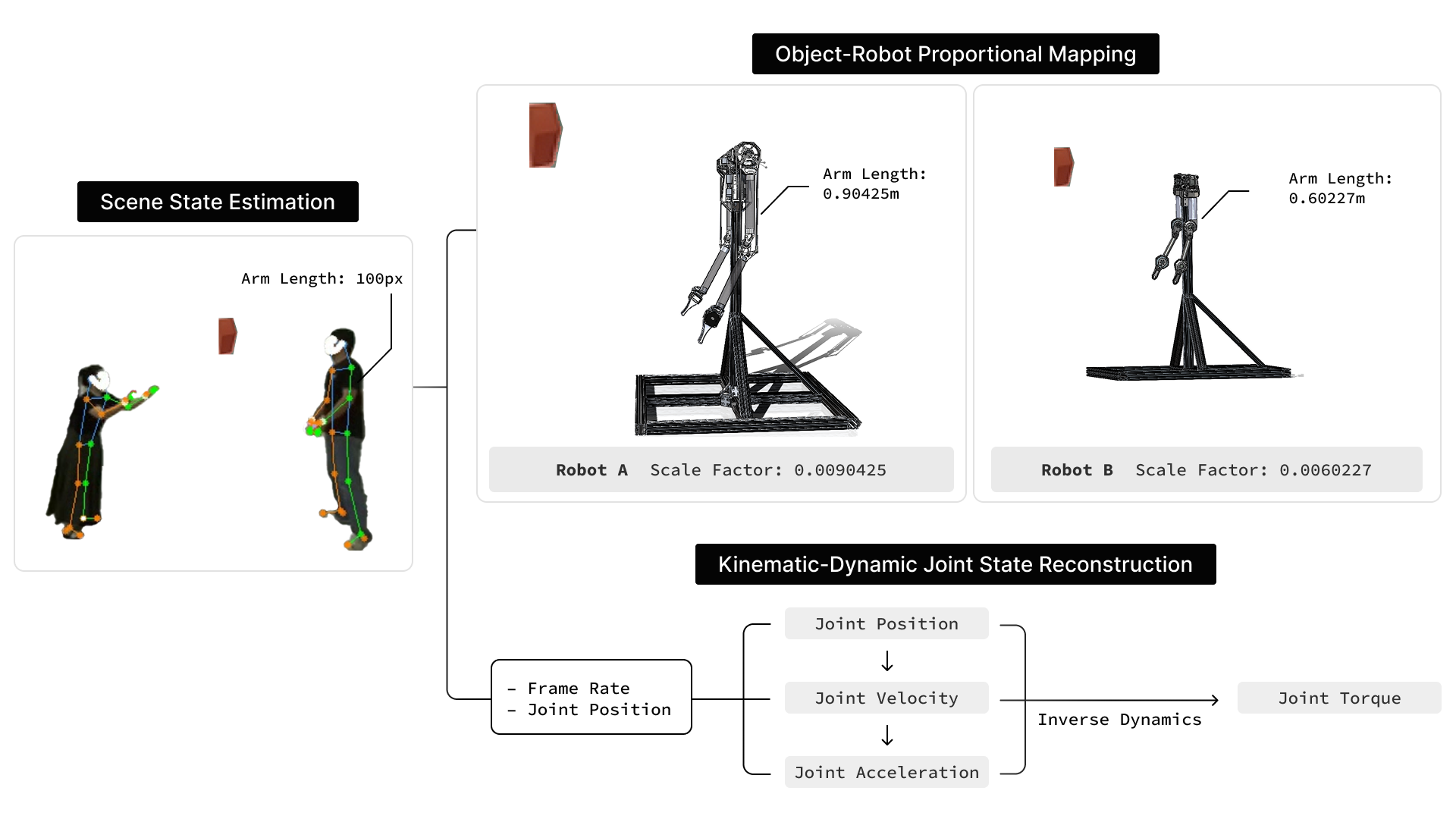}
    \caption{Robot Model-Based Data Regeneration Pipeline from Human Demonstration Videos. The pipeline consists of three main stages: (1) Scene State Estimation, which detects and tracks box objects and human pose from demonstration videos; (2) Object-Robot Proportional Mapping, which scales object dimensions and positions relative to the robot base frame; and (3) Kinematic-Dynamic Joint State Reconstruction, which maps human joint positions to robot configurations, differentiates to obtain velocities and accelerations, and computes required joint torques through inverse dynamics including external forces.}
    \label{fig:data_regen}
\end{figure}

A comprehensive framework is developed to regenerate robot training data from human demonstration videos, enabling the extraction of dynamic information without direct robot interaction. As illustrated in~\cref{fig:data_regen}, the approach consists of three main components: scene state estimation, object-robot proportional mapping, and kinematic-dynamic joint state reconstruction. This data regeneration pipeline systematically transforms human demonstrations into robot execution parameters while preserving essential dynamic characteristics.

\paragraph{Scene State Estimation}
Scene state estimation is performed to extract spatial information from video frames. This process involves detecting and tracking box objects to determine their positions and dimensions within the scene coordinate system. Additionally, human pose estimation is conducted to track the demonstrator's joint positions, providing the motion data necessary for the subsequent mapping process.

\paragraph{Object-Robot Proportional Mapping}
To adapt the human demonstration environment to the robot's workspace, object-robot proportional mapping is implemented. This process transforms detected objects' positions and dimensions to the robot's base coordinate frame:
\begin{equation}
{}^{R}\mathbf{p}_\mathrm{obj} = \mathbf{T}_{S}^{R} \cdot {}^{S}\mathbf{p}_\mathrm{obj}
\end{equation}
\begin{equation}
{}^{R}\mathbf{d}_\mathrm{obj} = s \cdot {}^{S}\mathbf{d}_\mathrm{obj}
\end{equation}
where ${}^{R}\mathbf{p}_\mathrm{obj}$ represents the object position expressed in robot frame $R$, ${}^{S}\mathbf{p}_\mathrm{obj}$ represents the object position expressed in scene frame $S$, $\mathbf{T}_{S}^{R}$ is the transformation matrix from scene frame to robot frame, ${}^{R}\mathbf{d}_\mathrm{obj}$ and ${}^{S}\mathbf{d}_\mathrm{obj}$ denote the object dimensions expressed in the robot and scene frames respectively, and $s$ is the scaling factor determined by the ratio between the robot arm length obtained from the robot model and the human arm length measured in pixels from the video.

\paragraph{Kinematic-Dynamic Joint State Reconstruction}
The robot's kinematic and dynamic states are reconstructed from the human demonstration. The joint positions are obtained through direct joint position mapping between human and robot:
\begin{equation}
\mathbf{q} = f_\mathrm{map}(\mathbf{q}_\mathrm{human})
\end{equation}
where $\mathbf{q}$ represents the robot joint positions and $f_\mathrm{map}$ is the mapping function that accounts for the different kinematic structures between human and robot.

Joint velocities and accelerations are calculated by differentiating the joint positions with respect to time using the video's frame rate:
\begin{equation}
\dot{\mathbf{q}} = \frac{d\mathbf{q}}{dt} \approx \frac{\mathbf{q}_{t+\Delta t} - \mathbf{q}_t}{\Delta t}
\end{equation}
\begin{equation}
\ddot{\mathbf{q}} = \frac{d^2\mathbf{q}}{dt^2} \approx \frac{\dot{\mathbf{q}}_{t+\Delta t} - \dot{\mathbf{q}}_t}{\Delta t}
\end{equation}
where $\Delta t$ is the time interval between consecutive frames.

With the complete kinematic state ($\mathbf{q}$, $\dot{\mathbf{q}}$, $\ddot{\mathbf{q}}$), inverse dynamics is employed to compute the required joint torques for task execution:
\begin{equation}
\boldsymbol{\tau} = \mathbf{M}(\mathbf{q})\ddot{\mathbf{q}} + \mathbf{C}(\mathbf{q},\dot{\mathbf{q}})\dot{\mathbf{q}} + \mathbf{G}(\mathbf{q}) + \mathbf{J}^T(\mathbf{q})\mathbf{F}_\mathrm{ext}
\end{equation}
where $\mathbf{M}(\mathbf{q})$ is the inertia matrix, $\mathbf{C}(\mathbf{q},\dot{\mathbf{q}})$ accounts for Coriolis and centrifugal effects, $\mathbf{G}(\mathbf{q})$ represents gravitational forces, $\mathbf{J}(\mathbf{q})$ is the Jacobian matrix, and $\mathbf{F}_\mathrm{ext}$ denotes external forces encountered during interaction with objects. The inclusion of external forces is crucial for accurately modeling tasks involving object manipulation and environmental contact.

This comprehensive data regeneration pipeline enables the transformation of human demonstrations into robot execution data with full kinematic and dynamic information, facilitating more efficient transfer of human skills to robotic systems.

\subsection{Latent Adaptive Planner (LAP)}
\label{sec:lap}
Our Latent Adaptive Planner (LAP) introduces a powerful framework for modeling and executing complex dynamic manipulation tasks based on human demonstrations.

\paragraph{Model}
Given a trajectory $\rvx = \{(o_t,a_t)\}_{t=1}^T$ consisting of observations and actions from demonstrations, LAP defines a joint probability distribution:
\begin{equation}
p_\theta(\rvx, \rvz) = p(\rvz) p_{\theta}(\rvx|\rvz)
\end{equation}
where $\rvz \in \mathbb{R}^d$ is the latent plan vector, $p(\rvz)$ is the prior model, and $p_{\theta}(\rvx|\rvz)$ is the trajectory generator. The prior model $p(\rvz)$ is an isotropic Gaussian $\rvz \sim \mathcal{N}(0, I_d)$.

The trajectory generator $p_{\theta}(\rvx|\rvz)$ is a conditional autoregressive model that produces actions based on the current state, historical context, and the latent plan vector:
\begin{equation}
p_{\theta}(\rvx|\rvz) = \prod_{t=1}^{T} p_{\theta}(\rvx_{(t)}|\rvx_{(t-K)}, ..., \rvx_{(t-1)}, \rvz)
\end{equation}
where $\rvx_{(t)} = (o_t, a_t)$, and $K$ is the context length. The observation $o_t$ includes box position, contact state, and previous timestep joint positions, velocities, and torques ($\mathbf{q}_{t-1}$, $\dot{\mathbf{q}}_{t-1}$, $\boldsymbol{\tau}_{t-1}$). Action $a_t$ consists of current timestep joint positions, velocities, and torques ($\mathbf{q}_t$, $\dot{\mathbf{q}}_t$, $\boldsymbol{\tau}_t$). $\rvx_{(0)}$ is a special learnable token. $\theta$ is implemented using a causal Transformer model~\citep{vaswani2017attention}. The latent plan vector $\rvz$ controls each step of action generation through cross-attention mechanisms as $p_\theta(a_{t}|o_{t-K:t-1},a_{t-K:t-1},\rvz)$. The action is assumed to follow a unimodal Gaussian distribution with fixed variance, i.e. $a_t\sim \mathcal{N}(g_\theta(o_{t-K:t},a_{t-K:t-1}, \rvz), I_{|a|})$.

\paragraph{Classical Variational Bayes Learning}
We employ classical variational Bayes (VB)~\cite{jordan1999introduction, blei2017variational, murphy2012machine,kong2025scalable} to learn LAP. Instead of learning an inference network (as in VAEs~\citep{kingma2013auto}), we directly optimize the latent plan vectors (local parameters in classical VB) for each trajectory using gradient descent. This iterative instance-level optimization offers greater precision in plan inference and greater flexibility that enables efficient adaptive (re)planning.

For each trajectory $\rvx$, we approximate the posterior distribution by $\mathcal{N}(\boldsymbol{\mu}, \boldsymbol{\sigma}^2)$, where $\boldsymbol{\mu}$ is the posterior mean vector and $\boldsymbol{\sigma}^2$ is the variance-covariance matrix, assumed to be diagonal for computational efficiency. These are local parameters specific to each trajectory, while $\theta$ are global parameters shared by all samples. LAP can be learned by maximizing the evidence lower bound (ELBO),
\begin{equation}
\label{eq:vl}
\begin{split}
    &\mathcal{L}(\theta, \boldsymbol{\mu}, \boldsymbol{\sigma})=  \E_{q(\rvz|\rvx)}[\log p_{\theta}(\rvx|\rvz) ] - \KL(q(\rvz|\rvx)\|p(\rvz)),
\end{split}
\end{equation}
where $\rvz\sim q(\rvz|\rvx)$ is sampled using reparameterization trick~\citep{kingma2013auto}.

The training procedure alternates between the optimization of local and global parameters: 
\begin{enumerate}[label=(\alph*), itemsep=0pt, topsep=0pt, partopsep=0pt]
    \item For each trajectory $\{\rvx_i\}_{i=1}^N$ in the mini-batch, optimize the corresponding local parameters $(\boldsymbol{\mu}_i, \boldsymbol{\sigma}_i)$ to maximize $\mathcal{L}(\theta, \boldsymbol{\mu}_i, \boldsymbol{\sigma}_i)$ with $T_\mathrm{local}$ steps of gradient descent (we set $T_\mathrm{local}=16$ in experiments). 
    \item Update the global parameters $\theta$ to maximize $\nicefrac{1}{N}\sum_{i=1}^N \mathcal{L}(\theta, \boldsymbol{\mu}_i, \boldsymbol{\sigma}_i)$.
\end{enumerate}
Detailed algorithms can be found in~\cref{appendix:algorithm}. 
This learning procedure also shares the similar intuition of fast-slow learning discussed in~\citep{kong2025scalable} with fast learning of local parameters and slower updates of global parameters.
This approach allows our model to efficiently learn an adaptive policy specific to manipulation scenarios while maintaining general knowledge about dynamics and control strategies across various tasks.

\paragraph{Variational Replanning for Test-time Adaptation}
We propose the variational replanning method that implements a principled Bayesian updating in the latent space, realizing the concept of \textit{planning as latent space inference}~\cite{kong2024latent}, which significantly improves the model efficiency and stability. 

With a learned LAP and initial observation $o_1$ during test time, we first sample $\rvz \sim p(\rvz|o_1)\propto p(\rvz)p(\rvx_{0:1}|\rvz)$ using $T_\mathrm{local}$ steps of gradient descent as the initial plan.
As new observations become available, we aim to adaptively update the latent plan $\rvz$ within the Bayesian framework. With replanning horizon $\Delta$, new observations $\rvx_{t+1:t+\Delta}$, and previously inferred posterior $q(\rvz|\rvx_{0:t})=\mathcal{N}(\boldsymbol{\mu}_t,\boldsymbol{\sigma}^2_t)$, we update our belief through Bayesian updating:
\begin{equation}
q(\rvz|\rvx_{0:t+\Delta})=\mathcal{N}(\boldsymbol{\mu}_{t+\Delta},\boldsymbol{\sigma}^2_{t+\Delta}) \propto q(\rvz|\rvx_{0:t})p(\rvx_{t+1:t+\Delta}|\rvx_{0:t}, \rvz)
\label{eq:update}
\end{equation}

Specifically, we use classical VB to optimize the local parameters $\boldsymbol{\mu}_{t+\Delta}$ and $\boldsymbol{\sigma}_{t+\Delta}$ as:
\begin{equation}
\boldsymbol{\mu}_{t+\Delta}, \boldsymbol{\sigma}_{t+\Delta} = \arg\max_{\boldsymbol{\mu}, \boldsymbol{\sigma}} \E_{q_{t+\Delta}}[\log p_\theta(\rvx_{t+1:t+\Delta}|\rvx_{0:t}, \rvz)] - \KL(q_{t+\Delta}\|q_t),
\label{eq:vr}
\end{equation}
where we define $q_{t+\Delta} \triangleq q(\rvz|\rvx_{0:t+\Delta})$ and $q_t \triangleq q(\rvz|\rvx_{0:t})$ for notational simplicity. This amortized computation enables efficient replanning with just $T_\mathrm{local}=1$ gradient step rather than the $16$ steps required for initial planning.

\begin{figure}[t!]
    \centering
    \includegraphics[width=\linewidth]{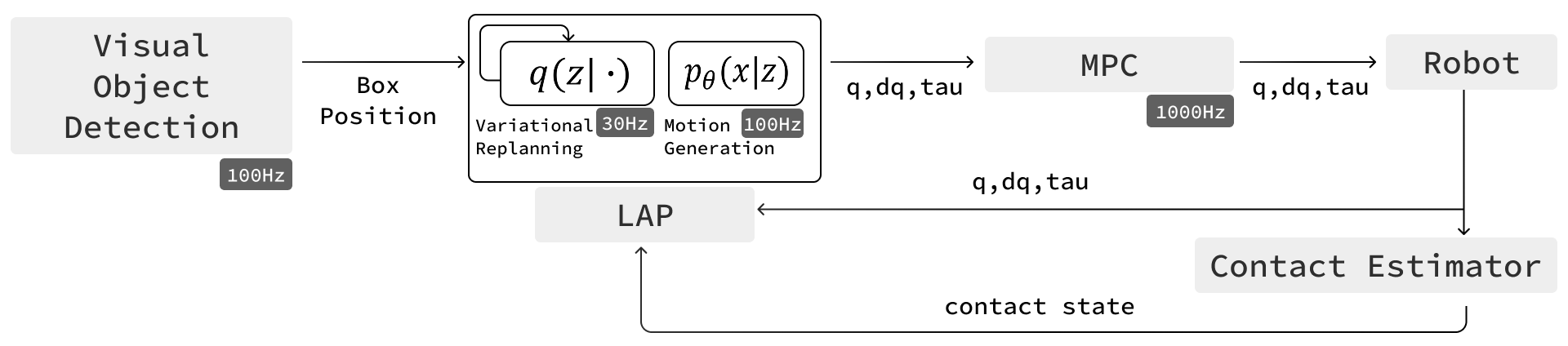}
    \vspace{-0.5cm}
    \caption{System Architecture for LAP Framework. The diagram illustrates our perception-planning-control pipeline. Camera input undergoes segmentation for object detection, providing box states to the Latent Adaptive Planner (LAP). LAP operates on a dual-rate hierarchy, performing updating of latent plan at 30Hz while generating motion commands at 100Hz. Reference joint positions, velocities, and torques from LAP are refined by Model Predictive Control before execution at the motor level. Joint states and contact state are fed back to the LAP for replanning. This architecture ensures smooth execution of dynamic nonprehensile manipulation tasks while respecting robot dynamics and physical constraints, enabling real-time adaptation to environmental changes.}
    \label{fig:system_architecture}
\end{figure}

The essence of our approach is treating the previously inferred distribution $q(\rvz|\rvx_{0:t})$ as the prior belief and updating it with a small number of gradient steps to obtain $q(\rvz|\rvx_{0:t+\Delta})$. As $t+\Delta \rightarrow T$ approaches the full trajectory length, the variational posterior $\mathcal{N}(\boldsymbol{\mu}_{t+\Delta},\boldsymbol{\sigma}^2_{t+\Delta})$ converges to the posterior distribution of $\mathcal{N}(\boldsymbol{\mu}_T,\boldsymbol{\sigma}^2_T)$ that would be inferred given the complete trajectory. This establishes theoretical consistency between our incremental replanning approach and the optimal plan that would be determined with complete information. Detailed algorithms can be found in~\cref{appendix:algorithm}. The integrated pipeline can be found in~\cref{fig:system_architecture}.
\paragraph{Discussion}
LAP with variational replanning offers a principled middle ground between traditional planning paradigms. Open-loop planning ($\rvz\sim p(\rvz|o_1)$ inferred once) provides computational efficiency but suffers from error accumulation over time. Closed-loop planning (resampling $\rvz\sim p(\rvz|\rvx_{0:t})$ at each step) allows adaptation but at a high computational cost. Our approach incrementally updates the latent plan through Bayesian inference, treating previous distributions as prior beliefs. This maintains the adaptive benefits of closed-loop planning while preserving computational efficiency, essentially performing online learning of the latent plan as new observations arrive.

Meanwhile, the objective in \cref{eq:vr} performs a KL-regularized proximal step around $q_t(\rvz)$: it increases the likelihood of the newly observed segment while staying close to $q_t$. Its optimizer is the exponential-tilting update in \cref{eq:update}, i.e., $q_t$ reweighted by the new-segment likelihood. With $q$ constrained to Gaussians, our single VB step implements a projected version of this tilt, acting as a small ``trust-region" move in latent space that yields stable, real-time adaptation.
\section{Experimental Results}
\label{sec:result}

We evaluate our LAP on a nonprehensile dynamic manipulation task of box catching, comparing its performance against several baseline methods including rule-based planning, behavior cloning, and Diffusion Policy approaches. This task represents a challenging example of nonprehensile dynamic manipulation, where the robot must initially interact with and control objects without stable grasping, using the dynamics of the arm movement to stabilize and then secure the incoming object.

\subsection{Experimental Setup}
\paragraph{Task Description}
Box catching constitutes the primary focus of this study, a challenging nonprehensile dynamic manipulation task where the robotic system must intercept, temporarily control, and securely grasp boxes of varying sizes, weights, and shapes handed or tossed by human operators. This task requires real-time trajectory adaptation, impact anticipation, and force modulation as the robot first manages dynamics using arm surfaces before transitioning to a stable grasp. 
\paragraph{Hardware Setup}
To evaluate generalizability, experiments were conducted on two distinct robot platforms with different sizes and arm configurations. The experimental procedure involved data regeneration processes tailored to each robot's specific characteristics, followed by training and validation phases. Both platforms were equipped with a ZED2 stereo camera that continuously tracks object position and trajectory through real-time segmentation algorithms. Human demonstrators introduced variability by throwing boxes with diverse velocities, heights, and release angles, ensuring comprehensive training and testing conditions. Detailed specifications of the robotic platforms are provided in the~\cref{appendix:robot_configurations}.

\paragraph{Baselines}We evaluated LAP against several baselines: (1) Model-based planner that determines optimal end-effector positions and arm configurations based on box dimensions and positions, using the inverse dynamics for superior tracking performance; (2) Transformer-based Behavior Cloning (BC) using a 3-layer decoder transformer (8 attention heads, batch size 12, hidden size 64, learning rate 2e-4, trained for 2500 epochs); (3) Diffusion Policy implemented with a 1D UNet architecture using 100 denoising steps based on the original repository~\citep{chi2023diffusion}. 
Our LAP uses identical architecture and hyper-parameters as BC with an additional cross-attention layer, where the number of $\rvz$ is 16, each with dimension 64, with $T_\mathrm{local}=16$ for plan optimization. All learning-based methods were trained on the same dataset regenerated from human demonstrations.

\begin{figure}[t!]
    \centering
    \includegraphics[height=0.23\textheight,width=\linewidth]{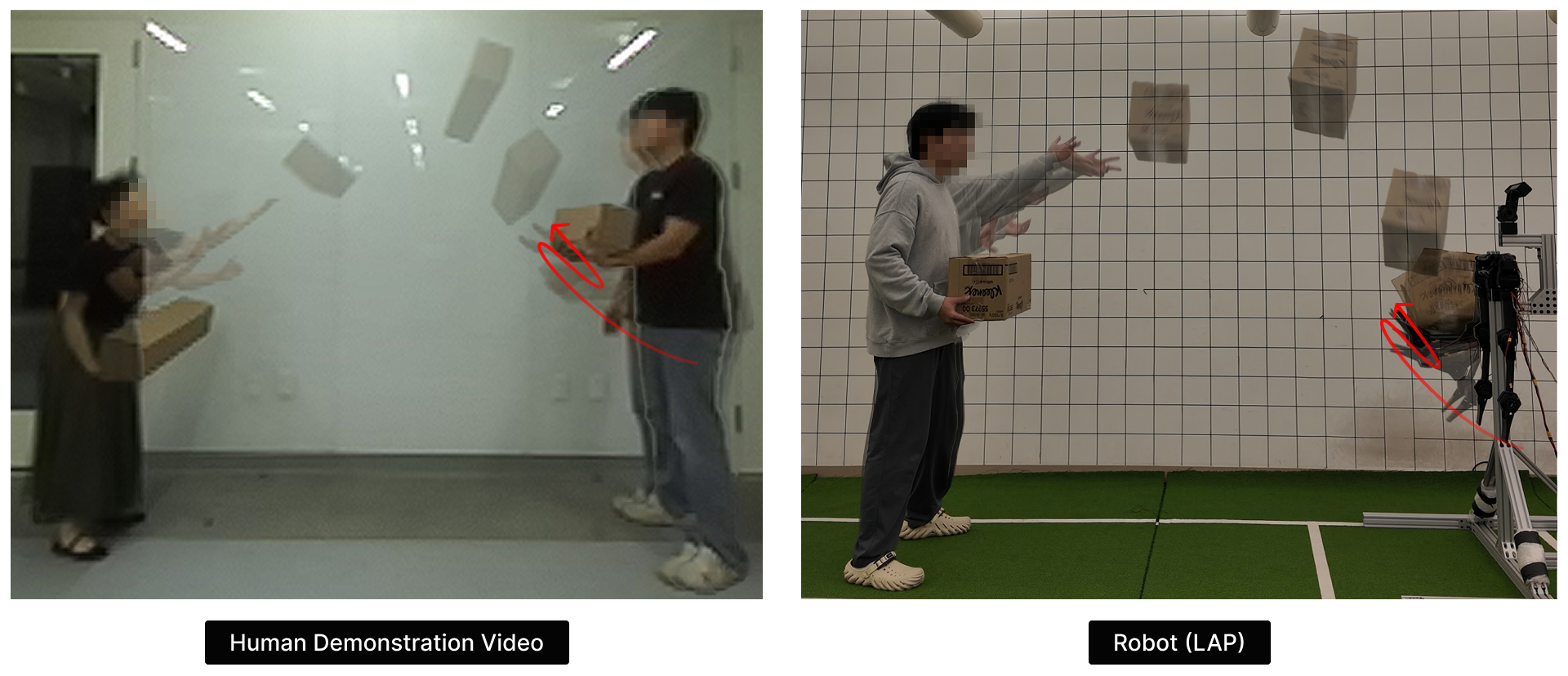}
    \vspace{-0.5cm}
    \caption{Impact-aware retreat trajectory for box catching motion learned with the Latent Adaptive Planner (LAP). The left panel shows a human demonstration where the subject absorbs impact by yielding their arm along the trajectory of the incoming box before returning to the nominal pose. The right panel shows the robot reproducing this compliant motion using LAP. The red curves indicate the impact-aware retreat trajectory that minimizes energy consumption during dynamic interaction.}
    \label{fig:real_robot}
\end{figure}

\begin{table}[t]
\scriptsize
\caption{Performance comparison of different methods on Robot A and B with various box types. Only trials where boxes were thrown within the reachable workspace of the fixed-base robots without excessive rotation were included in the success evaluation.}
\label{tab:performance_comparison}
\resizebox{\linewidth}{!}{
\centering
\setlength{\tabcolsep}{3pt}
\renewcommand{\arraystretch}{1.15}
\begin{tabular}{@{}c|c|cc|cc|cc@{}}
\hline
\textbf{Robot} & \textbf{Method} & \textbf{Success (count)} & \textbf{Energy (J)} & \textbf{Success (count)} & \textbf{Energy (J)} & \textbf{Success (count)} & \textbf{Energy (J)} \\
\hline
\multicolumn{2}{c|}{} & \multicolumn{2}{c|}{\textit{Box A (66×16.5×14cm, 453g)}} & \multicolumn{2}{c|}{\textit{Box B (61×30.5×30.5cm, 777g)}} & \multicolumn{2}{c}{\textit{Box C (67.1×38.2×23cm, 660g)}} \\
\hline
\multirow{4}{*}{\centering\textbf{Robot A}} & Model-based & $\mathbf{30/30}$ & $74.99$ & $\mathbf{30/30}$ & $57.31$ & $\mathbf{30/30}$ & $70.46$ \\
 & BC & $26/30$ & $31.39$ & $28/30$ & $38.50$ & $26/30$ & $37.40$ \\
 & Diffusion & $20/30$ & $42.53$ & $21/30$ & $45.21$ & $21/30$ & $44.86$ \\
 & LAP (Ours) & $29/30$ & $\mathbf{11.47}$ & $29/30$ & $\mathbf{12.12}$ & $\mathbf{30/30}$ & $\mathbf{11.41}$ \\
\hline
\multicolumn{2}{c|}{} & \multicolumn{2}{c|}{\textit{Box A (66×16.5×14cm, 453g)}} & \multicolumn{2}{c|}{\textit{Box D (48.3×22.9×24.8cm, 362g)}} & \multicolumn{2}{c}{\textit{Box E (49.5×12.7×23.5cm, 365g)}} \\
\hline
\multirow{4}{*}{\centering\textbf{Robot B}} & Model-based & $\mathbf{30/30}$ & $33.12$ & $\mathbf{30/30}$ & $33.17$ & $\mathbf{30/30}$ & $26.09$ \\
 & BC & $27/30$ & $15.82$ & $28/30$ & $15.40$ & $\mathbf{30/30}$ & $13.78$ \\
 & Diffusion & $24/30$ & $21.32$ & $22/30$ & $20.73$ & $23/30$ & $19.95$ \\
 & LAP (Ours) & $\mathbf{30/30}$ & $\mathbf{7.14}$ & $\mathbf{30/30}$ & $\mathbf{6.64}$ & $\mathbf{30/30}$ & $\mathbf{6.86}$ \\
\hline
\end{tabular}}
\end{table}

\subsection{Results and Analysis}

Our experiments demonstrate that LAP consistently outperforms the baselines in all metrics and box types on both robot platforms. \cref{tab:performance_comparison} summarizes the quantitative results for Robot A and Robot B, respectively. \cref{fig:real_robot} visualizes the impact-aware retreat trajectory for the the box catching motion learned by LAP using Robot B.

\paragraph{Model-based Planning}
The model-based approach achieved the highest success rate among all methods examined, however, this performance came at a significant energy cost due to its consistent application of high torque values for ensuring secure grasps. This conservative control strategy, while effective for task completion, proved suboptimal from an efficiency perspective. The implementation utilized an inverse-dynamic controller to enhance tracking performance and response time. The impedance control implementation offered inherent advantages in energy efficiency through its mass-spring-damper system, which naturally produces compliant motions when receiving impact forces. However, this approach presented substantial adaptability challenges, requiring manual gain tuning for variations in box weight and impact force, a significant limitation in dynamic environments.

\paragraph{Learning-based Methods}
Behavioral Cloning (BC) and Diffusion-based methods demonstrated some capacity to capture human-like manipulation behaviors, but exhibited lower success rates and less stable action outputs compared to other approaches. While integration with Model Predictive Control (MPC) improved motion smoothness, these methods still failed to achieve optimal performance across the evaluation metrics.

\paragraph{LAP Performance}
LAP balanced high success rates with superior energy efficiency compared to other policy-based methods. This performance can be attributed to two key factors: efficient replanning capabilities and effective encoding of contact dynamics and human-like retreat motions in the latent variable space during training. By learning to implement retreat motions through motion planning rather than relying solely on control-based implementations, LAP achieved a balanced performance profile. Though not matching the perfect success rates of the model-based approach with MPC, LAP demonstrated significantly improved energy efficiency through learned human-like motions that effectively absorbed impact forces and minimized overall energy consumption.


\section{Conclusion}
\label{sec:conclusion}
We introduced the Latent Adaptive Planner (LAP), a trajectory-level latent-variable policy that treats planning as inference and is learned from human demonstration videos via a robot-model-based data regeneration pipeline. During execution , LAP performs variational replanning by updating a posterior over the latent plan as new observations arrive, yielding real-time adaptation in dynamic scenes. This test-time adaptation behaves like a small trust-region move in latent space, providing stability without solving a full optimization at every step.

Our experiments demonstrate LAP's effectiveness across different manipulation scenarios and validate its transferability to diverse robot configurations. By regenerating appropriate training data from the same human demonstration videos for each platform, LAP successfully adapts to distinct kinematic and dynamic properties of different robots. This cross-platform success confirms that the combination of data regeneration and latent planning provides a robust framework for dynamic manipulation tasks regardless of robot embodiment, opening promising directions for scalable visuomotor learning from human demonstrations.
\newpage
\section*{Limitations}
\label{sec:limitations}
While our Latent Adaptive Planner demonstrates strong performance on dynamic manipulation tasks, several limitations remain and point to clear directions for future work.

\paragraph{Contact Modeling}
Our current approach employs a simplified binary contact representation and approximates contact locations at the wrist position. This simplification limits our ability to model the rich dynamics of multi-point contact interactions with varying force distributions, a critical aspect of dexterous manipulation where contact can occur across the entire arm surface with varying force distributions.

\paragraph{Kinematic Constraints}
To facilitate human-to-robot mapping, we restricted our demonstration analysis to three primary degrees of freedom (shoulder, elbow, and wrist pitch), excluding the full seven-DOF capability of human arms. This constraint, combined with our fixed-base manipulator assumption, prevents the system from leveraging whole-body coordination strategies that humans naturally employ during dynamic tasks, such as weight shifting and torso movements that contribute significantly to task performance and energy efficiency.

\paragraph{Perception Limitations}
Our reliance on 2D pose estimation from monocular video introduces fundamental constraints in spatial reasoning. The lack of depth information limits our ability to accurately reconstruct 3D trajectories and handle occlusions during dynamic movements, potentially affecting the fidelity of learned manipulation strategies in tasks requiring precise spatial awareness.

\paragraph{Prior Model Limitations}
From a generative modeling perspective, LAP employs a simple latent prior that may under-represent complex, multi-modal expert behaviors. More structured prior models, such as energy-based models~\citep{pang2020learning,pmlr-v216-kong23a,kong2021unsupervised,cheng2025latent}, diffusion models~\citep{yu2022latent}, and other expressive families~\citep{kong2024molecule,pmlr-v202-xu23j}, could  provide richer representations of the latent space, enabling more nuanced planning capabilities and better generalization to novel scenarios.

\paragraph{Future Work}
We aim to incorporate continuous contact models with multi-point force estimation, strengthen 3D perception for accurate spatial reasoning, extend to mobile platforms with whole-body coordination, and explore more expressive latent priors. These advances will be important for approaching human-level dexterity while maintaining the real-time performance required for dynamic manipulation.
\clearpage
\acknowledgments{Y.~W. was partially supported by NSF DMS-2015577, NSF DMS-2415226, and a gift fund from Amazon. We gratefully acknowledge Lambda, Inc. for providing compute resources for this project. We thank Sirui Xie for early collaboration and conceptual discussions.}


\bibliography{example}  

\begin{thebibliography}{59}
\providecommand{\natexlab}[1]{#1}
\providecommand{\url}[1]{\texttt{#1}}
\expandafter\ifx\csname urlstyle\endcsname\relax
  \providecommand{\doi}[1]{doi: #1}\else
  \providecommand{\doi}{doi: \begingroup \urlstyle{rm}\Url}\fi

\bibitem[Mason and Lynch(1993)]{mason1993dynamic}
M.~T. Mason and K.~M. Lynch.
\newblock Dynamic manipulation.
\newblock In \emph{Proceedings of 1993 IEEE/RSJ International Conference on Intelligent Robots and Systems (IROS'93)}, volume~1, pages 152--159. IEEE, 1993.

\bibitem[Ruggiero et~al.(2018)Ruggiero, Lippiello, and Siciliano]{ruggiero2018nonprehensile}
F.~Ruggiero, V.~Lippiello, and B.~Siciliano.
\newblock Nonprehensile dynamic manipulation: A survey.
\newblock \emph{IEEE Robotics and Automation Letters}, 3\penalty0 (3):\penalty0 1711--1718, 2018.

\bibitem[Noh et~al.(2021)Noh, Nam, Ahn, Chae, Lee, Gillespie, and Hong]{noh2021surface}
D.~Noh, H.~Nam, M.~S. Ahn, H.~Chae, S.~Lee, K.~Gillespie, and D.~Hong.
\newblock Surface material dataset for robotics applications (smdra): A dataset with friction coefficient and rgb-d for surface segmentation.
\newblock In \emph{2020 25th International Conference on Pattern Recognition (ICPR)}, pages 6275--6281. IEEE, 2021.

\bibitem[Noh(2024)]{noh2024control}
D.~Noh.
\newblock \emph{Control and Motion Planning for a Low-Inertia Multi-DoF Robotic Manipulator with Proprioceptive Actuators for Dynamic Manipulation}.
\newblock University of California, Los Angeles, 2024.

\bibitem[Lynch and Mason(1999)]{lynch1999dynamic}
K.~M. Lynch and M.~T. Mason.
\newblock Dynamic nonprehensile manipulation: Controllability, planning, and experiments.
\newblock \emph{The International Journal of Robotics Research}, 18\penalty0 (1):\penalty0 64--92, 1999.

\bibitem[Tsuji et~al.(2015)Tsuji, Ohkuma, and Sakaino]{tsuji2015dynamic}
T.~Tsuji, J.~Ohkuma, and S.~Sakaino.
\newblock Dynamic object manipulation considering contact condition of robot with tool.
\newblock \emph{IEEE Transactions on Industrial Electronics}, 63\penalty0 (3):\penalty0 1972--1980, 2015.

\bibitem[Gong et~al.(2023)Gong, Sun, Nair, Bidwai, CS, Grezmak, Sartoretti, and Daltorio]{gong2023legged}
Y.~Gong, G.~Sun, A.~Nair, A.~Bidwai, R.~CS, J.~Grezmak, G.~Sartoretti, and K.~A. Daltorio.
\newblock Legged robots for object manipulation: A review.
\newblock \emph{Frontiers in Mechanical Engineering}, 9:\penalty0 1142421, 2023.

\bibitem[Yu et~al.(2016)Yu, Bauza, Fazeli, and Rodriguez]{yu2016more}
K.-T. Yu, M.~Bauza, N.~Fazeli, and A.~Rodriguez.
\newblock More than a million ways to be pushed. a high-fidelity experimental dataset of planar pushing.
\newblock In \emph{2016 IEEE/RSJ international conference on intelligent robots and systems (IROS)}, pages 30--37. IEEE, 2016.

\bibitem[Fazeli et~al.(2017)Fazeli, Kolbert, Tedrake, and Rodriguez]{fazeli2017parameter}
N.~Fazeli, R.~Kolbert, R.~Tedrake, and A.~Rodriguez.
\newblock Parameter and contact force estimation of planar rigid-bodies undergoing frictional contact.
\newblock \emph{The International Journal of Robotics Research}, 36\penalty0 (13-14):\penalty0 1437--1454, 2017.

\bibitem[Zeng et~al.(2020)Zeng, Song, Lee, Rodriguez, and Funkhouser]{zeng2020tossingbot}
A.~Zeng, S.~Song, J.~Lee, A.~Rodriguez, and T.~Funkhouser.
\newblock Tossingbot: Learning to throw arbitrary objects with residual physics.
\newblock \emph{IEEE Transactions on Robotics}, 36\penalty0 (4):\penalty0 1307--1319, 2020.

\bibitem[Ha and Song(2022)]{ha2022flingbot}
H.~Ha and S.~Song.
\newblock Flingbot: The unreasonable effectiveness of dynamic manipulation for cloth unfolding.
\newblock In \emph{Conference on Robot Learning}, pages 24--33. PMLR, 2022.

\bibitem[Chi et~al.(2024)Chi, Burchfiel, Cousineau, Feng, and Song]{chi2024iterative}
C.~Chi, B.~Burchfiel, E.~Cousineau, S.~Feng, and S.~Song.
\newblock Iterative residual policy: for goal-conditioned dynamic manipulation of deformable objects.
\newblock \emph{The International Journal of Robotics Research}, 43\penalty0 (4):\penalty0 389--404, 2024.

\bibitem[Chi et~al.(2023)Chi, Feng, Du, Xu, Cousineau, Burchfiel, and Song]{chi2023diffusion}
C.~Chi, S.~Feng, Y.~Du, Z.~Xu, E.~Cousineau, B.~Burchfiel, and S.~Song.
\newblock Diffusion policy: Visuomotor policy learning via action diffusion.
\newblock \emph{Proceedings of Robotics: Science and Systems (RSS)}, 2023.

\bibitem[Lepert et~al.(2025)Lepert, Fang, and Bohg]{lepert2025phantom}
M.~Lepert, J.~Fang, and J.~Bohg.
\newblock Phantom: Training robots without robots using only human videos.
\newblock \emph{arXiv preprint arXiv:2503.00779}, 2025.

\bibitem[Shridhar et~al.(2022)Shridhar, Manuelli, and Fox]{shridhar2022cliport}
M.~Shridhar, L.~Manuelli, and D.~Fox.
\newblock Cliport: What and where pathways for robotic manipulation.
\newblock In \emph{Conference on robot learning}, pages 894--906. PMLR, 2022.

\bibitem[Mandlekar et~al.(2020)Mandlekar, Ramos, Boots, Savarese, Fei-Fei, Garg, and Fox]{mandlekar2020iris}
A.~Mandlekar, F.~Ramos, B.~Boots, S.~Savarese, L.~Fei-Fei, A.~Garg, and D.~Fox.
\newblock Iris: Implicit reinforcement without interaction at scale for learning control from offline robot manipulation data.
\newblock In \emph{2020 IEEE International Conference on Robotics and Automation (ICRA)}, pages 4414--4420. IEEE, 2020.

\bibitem[Zhang et~al.(2018)Zhang, McCarthy, Jow, Lee, Chen, Goldberg, and Abbeel]{zhang2018deep}
T.~Zhang, Z.~McCarthy, O.~Jow, D.~Lee, X.~Chen, K.~Goldberg, and P.~Abbeel.
\newblock Deep imitation learning for complex manipulation tasks from virtual reality teleoperation.
\newblock In \emph{2018 IEEE international conference on robotics and automation (ICRA)}, pages 5628--5635. Ieee, 2018.

\bibitem[Mandlekar et~al.(2019)Mandlekar, Booher, Spero, Tung, Gupta, Zhu, Garg, Savarese, and Fei-Fei]{mandlekar2019scaling}
A.~Mandlekar, J.~Booher, M.~Spero, A.~Tung, A.~Gupta, Y.~Zhu, A.~Garg, S.~Savarese, and L.~Fei-Fei.
\newblock Scaling robot supervision to hundreds of hours with roboturk: Robotic manipulation dataset through human reasoning and dexterity.
\newblock In \emph{2019 IEEE/RSJ International Conference on Intelligent Robots and Systems (IROS)}, pages 1048--1055. IEEE, 2019.

\bibitem[Rajeswaran et~al.(2017)Rajeswaran, Kumar, Gupta, Vezzani, Schulman, Todorov, and Levine]{rajeswaran2017learning}
A.~Rajeswaran, V.~Kumar, A.~Gupta, G.~Vezzani, J.~Schulman, E.~Todorov, and S.~Levine.
\newblock Learning complex dexterous manipulation with deep reinforcement learning and demonstrations.
\newblock \emph{arXiv preprint arXiv:1709.10087}, 2017.

\bibitem[Chen et~al.(2021)Chen, Lu, Rajeswaran, Lee, Grover, Laskin, Abbeel, Srinivas, and Mordatch]{chen2021decision}
L.~Chen, K.~Lu, A.~Rajeswaran, K.~Lee, A.~Grover, M.~Laskin, P.~Abbeel, A.~Srinivas, and I.~Mordatch.
\newblock Decision transformer: Reinforcement learning via sequence modeling.
\newblock In \emph{Advances in Neural Information Processing Systems}, volume~34, pages 15084--15097, 2021.

\bibitem[Kim et~al.(2014)Kim, Shukla, and Billard]{kim2014catching}
S.~Kim, A.~Shukla, and A.~Billard.
\newblock Catching objects in flight.
\newblock \emph{IEEE Transactions on Robotics}, 30\penalty0 (5):\penalty0 1049--1065, 2014.

\bibitem[Lampariello et~al.(2011)Lampariello, Nguyen-Tuong, Castellini, Hirzinger, and Peters]{lampariello2011trajectory}
R.~Lampariello, D.~Nguyen-Tuong, C.~Castellini, G.~Hirzinger, and J.~Peters.
\newblock Trajectory planning for optimal robot catching in real-time.
\newblock In \emph{2011 IEEE International Conference on Robotics and Automation}, pages 3719--3726. IEEE, 2011.

\bibitem[Yan et~al.(2024)Yan, Stouraitis, Moura, Xu, Gienger, and Vijayakumar]{yan2024impact}
L.~Yan, T.~Stouraitis, J.~Moura, W.~Xu, M.~Gienger, and S.~Vijayakumar.
\newblock Impact-aware bimanual catching of large-momentum objects.
\newblock \emph{IEEE Transactions on Robotics}, 2024.

\bibitem[Ebert et~al.(2021)Ebert, Yang, Schmeckpeper, Bucher, Georgakis, Daniilidis, Finn, and Levine]{ebert2021bridge}
F.~Ebert, Y.~Yang, K.~Schmeckpeper, B.~Bucher, G.~Georgakis, K.~Daniilidis, C.~Finn, and S.~Levine.
\newblock Bridge data: Boosting generalization of robotic skills with cross-domain datasets.
\newblock \emph{arXiv preprint arXiv:2109.13396}, 2021.

\bibitem[Sharma et~al.(2018)Sharma, Mohan, Pinto, and Gupta]{sharma2018multiple}
P.~Sharma, L.~Mohan, L.~Pinto, and A.~Gupta.
\newblock Multiple interactions made easy (mime): Large scale demonstrations data for imitation.
\newblock In \emph{Conference on robot learning}, pages 906--915. PMLR, 2018.

\bibitem[Song et~al.(2020)Song, Zeng, Lee, and Funkhouser]{song2020grasping}
S.~Song, A.~Zeng, J.~Lee, and T.~Funkhouser.
\newblock Grasping in the wild: Learning 6dof closed-loop grasping from low-cost demonstrations.
\newblock \emph{IEEE Robotics and Automation Letters}, 5\penalty0 (3):\penalty0 4978--4985, 2020.

\bibitem[Mandikal and Grauman(2022)]{mandikal2022dexvip}
P.~Mandikal and K.~Grauman.
\newblock Dexvip: Learning dexterous grasping with human hand pose priors from video.
\newblock In \emph{Conference on Robot Learning}, pages 651--661. PMLR, 2022.

\bibitem[Cai et~al.(2016)Cai, Kitani, and Sato]{cai2016understanding}
M.~Cai, K.~M. Kitani, and Y.~Sato.
\newblock Understanding hand-object manipulation with grasp types and object attributes.
\newblock In \emph{Robotics: science and systems}, volume~3, 2016.

\bibitem[Chen et~al.(2021)Chen, Tu, Kang, Chen, Bao, Zhang, and Yuan]{chen2021joint}
Y.~Chen, Z.~Tu, D.~Kang, R.~Chen, L.~Bao, Z.~Zhang, and J.~Yuan.
\newblock Joint hand-object 3d reconstruction from a single image with cross-branch feature fusion.
\newblock \emph{IEEE Transactions on Image Processing}, 30:\penalty0 4008--4021, 2021.

\bibitem[Young et~al.(2021)Young, Gandhi, Tulsiani, Gupta, Abbeel, and Pinto]{young2021visual}
S.~Young, D.~Gandhi, S.~Tulsiani, A.~Gupta, P.~Abbeel, and L.~Pinto.
\newblock Visual imitation made easy.
\newblock In \emph{Conference on Robot learning}, pages 1992--2005. PMLR, 2021.

\bibitem[Schmeckpeper et~al.(2020{\natexlab{a}})Schmeckpeper, Xie, Rybkin, Tian, Daniilidis, Levine, and Finn]{schmeckpeper2020learning}
K.~Schmeckpeper, A.~Xie, O.~Rybkin, S.~Tian, K.~Daniilidis, S.~Levine, and C.~Finn.
\newblock Learning predictive models from observation and interaction.
\newblock In \emph{European Conference on Computer Vision}, pages 708--725. Springer, 2020{\natexlab{a}}.

\bibitem[Schmeckpeper et~al.(2020{\natexlab{b}})Schmeckpeper, Rybkin, Daniilidis, Levine, and Finn]{schmeckpeper2020reinforcement}
K.~Schmeckpeper, O.~Rybkin, K.~Daniilidis, S.~Levine, and C.~Finn.
\newblock Reinforcement learning with videos: Combining offline observations with interaction.
\newblock \emph{arXiv preprint arXiv:2011.06507}, 2020{\natexlab{b}}.

\bibitem[Chen et~al.(2024)Chen, Xu, Dharmarajan, Irshad, Cheng, Keutzer, Tomizuka, Vuong, and Goldberg]{chen2024rovi}
L.~Y. Chen, C.~Xu, K.~Dharmarajan, M.~Z. Irshad, R.~Cheng, K.~Keutzer, M.~Tomizuka, Q.~Vuong, and K.~Goldberg.
\newblock Rovi-aug: Robot and viewpoint augmentation for cross-embodiment robot learning.
\newblock \emph{arXiv preprint arXiv:2409.03403}, 2024.

\bibitem[Sharma et~al.(2019)Sharma, Pathak, and Gupta]{EgoMimic}
P.~Sharma, D.~Pathak, and A.~Gupta.
\newblock Egomimic: Imitation learning for robotic manipulation from third-person videos.
\newblock In \emph{Conference on Neural Information Processing Systems (NeurIPS)}, 2019.

\bibitem[Lepert et~al.(2023)Lepert, Kumar, Zhan, Wang, Dasari, and Levine]{Phantom}
M.~Lepert, V.~Kumar, A.~Zhan, H.~Wang, S.~Dasari, and S.~Levine.
\newblock Phantom: Learning generalizable mobile manipulation from human videos with motion guidance.
\newblock In \emph{Conference on Robot Learning (CoRL)}, 2023.

\bibitem[Khazatsky et~al.(2024)Khazatsky, Pertsch, Nair, Balakrishna, Dasari, Karamcheti, Nasiriany, Srirama, Chen, Ellis, et~al.]{DROID}
A.~Khazatsky, K.~Pertsch, S.~Nair, A.~Balakrishna, S.~Dasari, S.~Karamcheti, S.~Nasiriany, M.~K. Srirama, L.~Y. Chen, K.~Ellis, et~al.
\newblock Droid: A large-scale in-the-wild robot manipulation dataset.
\newblock In \emph{arXiv preprint arXiv:2403.12945}, 2024.

\bibitem[Qin et~al.(2022)Qin, Wu, Liu, Jiang, Yang, Fu, and Wang]{DexMV}
Y.~Qin, Y.-H. Wu, S.~Liu, H.~Jiang, R.~Yang, Y.~Fu, and X.~Wang.
\newblock Dexmv: Imitation learning for dexterous manipulation from human videos.
\newblock In \emph{Robotics: Science and Systems (RSS)}, 2022.

\bibitem[Schaal(1999)]{schaal1999imitation}
S.~Schaal.
\newblock Is imitation learning the route to humanoid robots?
\newblock \emph{Trends in cognitive sciences}, 3\penalty0 (6):\penalty0 233--242, 1999.

\bibitem[Argall et~al.(2009)Argall, Chernova, Veloso, and Browning]{argall2009survey}
B.~D. Argall, S.~Chernova, M.~Veloso, and B.~Browning.
\newblock A survey of robot learning from demonstration.
\newblock \emph{Robotics and autonomous systems}, 57\penalty0 (5):\penalty0 469--483, 2009.

\bibitem[Janner et~al.(2021)Janner, Li, and Levine]{janner2021offline}
M.~Janner, Q.~Li, and S.~Levine.
\newblock Offline reinforcement learning as one big sequence modeling problem.
\newblock In \emph{Advances in Neural Information Processing Systems}, volume~34, pages 1273--1286, 2021.

\bibitem[Janner et~al.(2022)Janner, Du, Tenenbaum, and Levine]{janner2022planning}
M.~Janner, Y.~Du, J.~B. Tenenbaum, and S.~Levine.
\newblock Planning with diffusion for flexible behavior synthesis.
\newblock \emph{arXiv preprint arXiv:2205.09991}, 2022.

\bibitem[Yang et~al.(2022)Yang, Schuurmans, Abbeel, and Nachum]{yang2022dichotomy}
S.~Yang, D.~Schuurmans, P.~Abbeel, and O.~Nachum.
\newblock Dichotomy of control: Separating what you can control from what you cannot.
\newblock In \emph{International Conference on Learning Representations (ICLR)}, 2022.

\bibitem[Paster et~al.(2022)Paster, McIlraith, and Ba]{paster2022you}
K.~Paster, S.~McIlraith, and J.~Ba.
\newblock You can't count on luck: Why decision transformers and rvs fail in stochastic environments.
\newblock In \emph{Advances in Neural Information Processing Systems}, volume~35, pages 38966--38979, 2022.

\bibitem[Kong et~al.(2024)Kong, Xu, Zhao, Pang, Xie, Lizarraga, Huang, Xie, and Wu]{kong2024latent}
D.~Kong, D.~Xu, M.~Zhao, B.~Pang, J.~Xie, A.~Lizarraga, Y.~Huang, S.~Xie, and Y.~N. Wu.
\newblock Latent plan transformer for trajectory abstraction: Planning as latent space inference.
\newblock In \emph{Conference on Neural Information Processing Systems}, 2024.

\bibitem[Ajay et~al.(2021)Ajay, Kumar, Agrawal, Levine, and Nachum]{ajay2021opal}
A.~Ajay, A.~Kumar, P.~Agrawal, S.~Levine, and O.~Nachum.
\newblock Opal: Offline primitive discovery for accelerating offline reinforcement learning.
\newblock In \emph{International Conference on Learning Representations}, 2021.

\bibitem[Lynch et~al.(2020)Lynch, Khansari, Xiao, Kumar, Tompson, Levine, and Sermanet]{lynch2020learning}
C.~Lynch, M.~Khansari, T.~Xiao, V.~Kumar, J.~Tompson, S.~Levine, and P.~Sermanet.
\newblock Learning latent plans from play.
\newblock In \emph{Conference on robot learning (CoRL)}, pages 1113--1132, 2020.

\bibitem[Kingma and Welling(2013)]{kingma2013auto}
D.~P. Kingma and M.~Welling.
\newblock Auto-encoding variational bayes.
\newblock \emph{arXiv preprint arXiv:1312.6114}, 2013.

\bibitem[Jordan et~al.(1999)Jordan, Ghahramani, Jaakkola, and Saul]{jordan1999introduction}
M.~I. Jordan, Z.~Ghahramani, T.~S. Jaakkola, and L.~K. Saul.
\newblock An introduction to variational methods for graphical models.
\newblock \emph{Machine learning}, 37\penalty0 (2):\penalty0 183--233, 1999.

\bibitem[Blei et~al.(2017)Blei, Kucukelbir, and McAuliffe]{blei2017variational}
D.~M. Blei, A.~Kucukelbir, and J.~D. McAuliffe.
\newblock Variational inference: A review for statisticians.
\newblock \emph{Journal of the American Statistical Association}, 112\penalty0 (518):\penalty0 859--877, 2017.

\bibitem[Vaswani et~al.(2017)Vaswani, Shazeer, Parmar, Uszkoreit, Jones, Gomez, Kaiser, and Polosukhin]{vaswani2017attention}
A.~Vaswani, N.~Shazeer, N.~Parmar, J.~Uszkoreit, L.~Jones, A.~N. Gomez, {\L}.~Kaiser, and I.~Polosukhin.
\newblock Attention is all you need.
\newblock In \emph{Advances in Neural Information Processing Systems}, volume~30, pages 5998--6008, 2017.

\bibitem[Murphy(2012)]{murphy2012machine}
K.~P. Murphy.
\newblock \emph{Machine Learning: A Probabilistic Perspective}.
\newblock Adaptive Computation and Machine Learning series. MIT Press, Cambridge, MA, 2012.

\bibitem[Kong et~al.(2025)Kong, Zhao, Xu, Pang, Wang, Honig, Si, Li, Xie, Xie, and Wu]{kong2025scalable}
D.~Kong, M.~Zhao, D.~Xu, B.~Pang, S.~Wang, E.~Honig, Z.~Si, C.~Li, J.~Xie, S.~Xie, and Y.~N. Wu.
\newblock Latent thought models with variational bayes inference-time computation.
\newblock In \emph{Forty-second International Conference on Machine Learning}, 2025.

\bibitem[Pang et~al.(2020)Pang, Han, Nijkamp, Zhu, and Wu]{pang2020learning}
B.~Pang, T.~Han, E.~Nijkamp, S.-C. Zhu, and Y.~N. Wu.
\newblock Learning latent space energy-based prior model.
\newblock \emph{Advances in Neural Information Processing Systems}, 33:\penalty0 21994--22008, 2020.

\bibitem[Kong et~al.(2023)Kong, Pang, Han, and Wu]{pmlr-v216-kong23a}
D.~Kong, B.~Pang, T.~Han, and Y.~N. Wu.
\newblock Molecule design by latent space energy-based modeling and gradual distribution shifting.
\newblock In R.~J. Evans and I.~Shpitser, editors, \emph{Proceedings of the Thirty-Ninth Conference on Uncertainty in Artificial Intelligence}, volume 216 of \emph{Proceedings of Machine Learning Research}, pages 1109--1120. PMLR, 31 Jul--04 Aug 2023.

\bibitem[Kong et~al.(2021)Kong, Pang, and Wu]{kong2021unsupervised}
D.~Kong, B.~Pang, and Y.~N. Wu.
\newblock Unsupervised meta-learning via latent space energy-based model of symbol vector coupling.
\newblock In \emph{Fifth Workshop on Meta-Learning at the Conference on Neural Information Processing Systems}, 2021.

\bibitem[Cheng et~al.(2025)Cheng, Kong, Xie, Lee, Wu, and Yang]{cheng2025latent}
S.~Cheng, D.~Kong, J.~Xie, K.~Lee, Y.~N. Wu, and Y.~Yang.
\newblock Latent space energy-based neural {ODE}s.
\newblock \emph{Transactions on Machine Learning Research}, 2025.
\newblock ISSN 2835-8856.

\bibitem[Yu et~al.(2022)Yu, Xie, Ma, Jia, Pang, Gao, Zhu, Zhu, and Wu]{yu2022latent}
P.~Yu, S.~Xie, X.~Ma, B.~Jia, B.~Pang, R.~Gao, Y.~Zhu, S.-C. Zhu, and Y.~N. Wu.
\newblock Latent diffusion energy-based model for interpretable text modeling.
\newblock \emph{arXiv preprint arXiv:2206.05895}, 2022.

\bibitem[Kong et~al.(2024)Kong, Huang, Xie, Honig, Xu, Xue, Lin, Zhou, Zhong, Zheng, and Wu]{kong2024molecule}
D.~Kong, Y.~Huang, J.~Xie, E.~Honig, M.~Xu, S.~Xue, P.~Lin, S.~Zhou, S.~Zhong, N.~Zheng, and Y.~N. Wu.
\newblock Molecule design by latent prompt transformer.
\newblock In \emph{The Thirty-eighth Annual Conference on Neural Information Processing Systems}, 2024.

\bibitem[Xu et~al.(2023)Xu, Kong, Xu, Ji, Pang, Fung, and Wu]{pmlr-v202-xu23j}
Y.~Xu, D.~Kong, D.~Xu, Z.~Ji, B.~Pang, P.~Fung, and Y.~N. Wu.
\newblock Diverse and faithful knowledge-grounded dialogue generation via sequential posterior inference.
\newblock In A.~Krause, E.~Brunskill, K.~Cho, B.~Engelhardt, S.~Sabato, and J.~Scarlett, editors, \emph{Proceedings of the 40th International Conference on Machine Learning}, volume 202 of \emph{Proceedings of Machine Learning Research}, pages 38518--38534. PMLR, 23--29 Jul 2023.

\end{thebibliography}

\newpage
\appendix
\section{Model Predictive Control (MPC)}
\label{appendix:mpc}

Dynamic Model Predictive Control (MPC) provides an effective framework for safely tracking joint-space references 
$\hat{q}_k, \hat{\dot{q}}_k, \hat{\tau}_k$ generated by learning policies. 
When implemented, this approach significantly mitigates the inherently noisy characteristics of learning-based control policies 
while maintaining high tracking performance. By incorporating the robot's dynamics and physical constraints into the optimization problem, 
the controller ensures smooth and secure motions even when the underlying learning policy produces potentially suboptimal commands. 

The formulated cost function penalizes deviations between desired and actual joint states while simultaneously limiting control effort, 
allowing the robot to operate efficiently within specified joint position, velocity, and torque limits. 
This balance between precise reference tracking and constraint satisfaction ultimately leads to more natural robot movements, 
enhancing the overall execution quality of learned behaviors while preventing excessive energy consumption and 
vibrational behaviors that might otherwise occur with direct policy execution.

\paragraph{QP Formulation} 
Define the robot state at time step $k$ as
\[
\mathbf{x}_k = 
\begin{bmatrix}
\mathbf{q}_k \\
\dot{\mathbf{q}}_k
\end{bmatrix},
\]
where $\mathbf{q}_k$ and $\dot{\mathbf{q}}_k$ represent the joint angles and joint velocities, respectively. Let $\mathbf{u}_k$ be the control input (torque). 
Over a prediction horizon of length $N_p$, the following cost function is minimized:
\[
\sum_{k=i}^{i+N_p} 
\Big[
(\hat{\mathbf{q}}_k - \mathbf{q}_k)^T \, Q_q \, (\hat{\mathbf{q}}_k - \mathbf{q}_k)
\;+\;
(\hat{\dot{\mathbf{q}}}_k - \dot{\mathbf{q}}_k)^T \, Q_{\dot{q}} \, (\hat{\dot{\mathbf{q}}}_k - \dot{\mathbf{q}}_k)
\;+\;
\mathbf{u}_k^T \, Q_u \, \mathbf{u}_k
\Big],
\]
subject to the robot dynamics and constraints. The linearized (or identified) system model is given by
\[
\mathbf{x}_{k+1} = A_k \, \mathbf{x}_k + B_k \, \mathbf{u}_k + \mathbf{r}_k,
\]
where $A_k$, $B_k$, and $\mathbf{r}_k$ approximate the robot's dynamics around a nominal operating point. The following constraints are imposed to ensure feasibility and safety:
\[
\begin{aligned}
\mathbf{u}_{\min} &\leq \mathbf{u}_k \leq \mathbf{u}_{\max}, \\
\mathbf{q}_{\min} &\leq \mathbf{q}_k \leq \mathbf{q}_{\max}, \\
\dot{\mathbf{q}}_{\min} &\leq \dot{\mathbf{q}}_k \leq \dot{\mathbf{q}}_{\max}, \\
\ddot{\mathbf{q}}_{\min} &\leq \ddot{\mathbf{q}}_k \leq \ddot{\mathbf{q}}_{\max}, \\
\mathbf{q}_0 &= \mathbf{q}_{\text{current}}, \quad
\dot{\mathbf{q}}_0 = \dot{\mathbf{q}}_{\text{current}},
\end{aligned}
\]
where $\mathbf{q}_{\min}, \mathbf{q}_{\max}, \dot{\mathbf{q}}_{\min}, \dot{\mathbf{q}}_{\max}$, and so forth denote predefined joint and actuator limits. 
Collecting the state and input variables into a decision vector 
\[
\mathbf{z}_k = 
\begin{bmatrix}
\mathbf{x}_k \\
\mathbf{u}_k
\end{bmatrix},
\]
the QP can be expressed in a compact form as
\[
\begin{aligned}
\min_{\{\mathbf{z}_k\}_{k=i}^{i+N_p}} \quad 
& \sum_{k=i}^{i+N_p} 
\bigl(
\mathbf{z}_k^T W_k \, \mathbf{z}_k + 2\, \mathbf{w}_k^T \, \mathbf{z}_k
\bigr), \\[4pt]
\text{subject to}\quad 
& \mathbf{x}_{k+1} = A_k \, \mathbf{x}_k + B_k \, \mathbf{u}_k + \mathbf{r}_k, \\
& \mathbf{u}_{\min} \le \mathbf{u}_k \le \mathbf{u}_{\max}, \\
& \mathbf{q}_{\min} \le \mathbf{q}_k \le \mathbf{q}_{\max}, \\
& \dot{\mathbf{q}}_{\min} \le \dot{\mathbf{q}}_k \le \dot{\mathbf{q}}_{\max}, \\
& \ddot{\mathbf{q}}_{\min} \le \ddot{\mathbf{q}}_k \le \ddot{\mathbf{q}}_{\max}, \\
& \mathbf{q}_0 = \mathbf{q}_{\text{current}}, \quad 
  \dot{\mathbf{q}}_0 = \dot{\mathbf{q}}_{\text{current}}.
\end{aligned}
\]
The matrices $W_k$ and vectors $\mathbf{w}_k$ encapsulate the quadratic and linear terms derived from expanding the cost function and incorporating the linearized dynamics. Solving this QP at each time step allows the robot to effectively track the learning policy's references $\hat{\mathbf{q}}_k, \hat{\dot{\mathbf{q}}}_k$ while honoring physical constraints and preserving safe operation.

\section{Robot Configurations}
\label{appendix:robot_configurations}

\subsection{Actuators}
\begin{table}[h!]
\centering
\caption{Specifications of Different Actuator Models}
\label{tab:actuator_specs}
\renewcommand{\arraystretch}{1.2}
\footnotesize
\begin{tabular}{lcccc}
\hline
\textbf{Specification} & \textbf{Koala BEAR} & \textbf{Koala BEAR Muscle} & \textbf{Panda BEAR} & \textbf{Panda BEAR Plus} \\
\hline
Dimensions (mm) & $63.5 \times 62 \times 37$ & $75 \times 67 \times 37.5$ & $113 \times 113 \times 49.7$ & $113 \times 113 \times 49.7$ \\
Weight (g) & 250 & 285 & 650 & 925 \\
Peak Torque (15 sec) & 4.2 Nm & 8 Nm & 16.8 Nm & 33 Nm \\
Peak Torque (1.5 sec) & 10.5 Nm & 20 Nm & 33.5 Nm & 67 Nm \\
\hline
\end{tabular}
\end{table}

\begin{table}[h!]
\centering
\caption{Actuator Configuration for Robot A}
\renewcommand{\arraystretch}{1.2}
\label{tab:robot_a_config}
\begin{tabular}{lc}
\hline
\textbf{Joint} & \textbf{Actuator Model} \\
\hline
Body & Panda BEAR Plus \\
Shoulder Pitch & Panda BEAR Plus \\
Shoulder Yaw & Panda BEAR \\
Elbow Pitch & Panda BEAR Plus \\
Wrist Pitch & Koala BEAR \\
Wrist Roll & Koala BEAR \\
\hline
\end{tabular}
\end{table}

\begin{table}[h!]
\centering
\caption{Actuator Configuration for Robot B}
\label{tab:robot_b_config}
\renewcommand{\arraystretch}{1.2}
\begin{tabular}{lc}
\hline
\textbf{Joint} & \textbf{Actuator Model} \\
\hline
Body & Koala BEAR Muscle \\
Shoulder Pitch & Koala BEAR Muscle \\
Shoulder Yaw & Koala BEAR Muscle \\
Elbow Pitch & Koala BEAR Muscle \\
Wrist Pitch & Koala BEAR Muscle \\
Wrist Roll & Koala BEAR Muscle \\
\hline
\end{tabular}
\end{table}

\subsection{Robot A}
\begin{figure}[h!]
    \centering
    \includegraphics[width=0.6\linewidth]{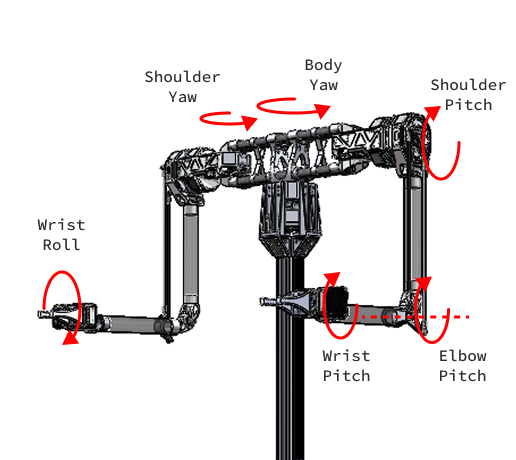}
    \caption{Coordinate system of Robot A}
    \label{fig:robot_a}
\end{figure}

\begin{table}[h!]
\centering
\caption{Robot A Configuration}
\label{tab:robot_A_config}
\renewcommand{\arraystretch}{1.2}
\renewcommand{\arraystretch}{1.2}
\begin{tabular}{lccc}
\hline
\textbf{Joint} & \textbf{Frame Coordination} & \textbf{Coordinate (m)} & \textbf{Joint Limit (rad)} \\
\hline
Body Yaw & $z$-axis rotation & (0.0, 0.0, 1.1276) & $[-1.57, 1.57]$ \\
Shoulder Yaw & $z$-axis rotation & (0.0, $\pm$0.21, 0.117) & $[-2.27, 2.27]$ \\
Shoulder Pitch & $y$-axis rotation & (0.0, $\pm$0.15, 0.0) & $[-1.57, 1.57]$ \\
Elbow Pitch & $y$-axis rotation & (0.0, 0.4, 0.0) & $[-1.57, 2.53]$ \\
Wrist Pitch & $y$-axis rotation & (0.0, 0.375, 0.0) & $[-2.44, 2.44]$ \\
Wrist Roll & $x$-axis rotation & (0.01925, 0.0, 0.0) & $[-1.57, 1.57]$ \\
\hline
\end{tabular}
\end{table}

\newpage
\subsection{Robot B}

\begin{table}[h!]
\centering
\caption{Robot B Configuration}
\label{tab:robot_B_config}
\renewcommand{\arraystretch}{1.2}
\renewcommand{\arraystretch}{1.2}
\begin{tabular}{lccc}
\hline
\textbf{Joint} & \textbf{Frame Coordination} & \textbf{Coordinate (m)} & \textbf{Joint Limit (rad)} \\
\hline
Body Yaw & $z$-axis rotation & (0.0, 0.0, 1.1327) & $[-1.57, 1.57]$ \\
Shoulder Pitch & $y$-axis rotation & (0.0, $\pm$0.19475, 0.0) & $[-1.57, 1.57]$ \\
Shoulder Roll & $x$-axis rotation & (0.0, 0.0, 0.0) & $\begin{array}{l}[-1.57, 0.25] \end{array}$ \\
Elbow Yaw & $z$-axis rotation & (0.0, 0.0, 0.0) & $\begin{array}{l}[-0.79, 1.57] \end{array}$ \\
Elbow Pitch & $y$-axis rotation & (0.02, 0.0, -0.27585) & $[-0.3, 2.9]$ \\
Wrist Pitch & $y$-axis rotation & (0.23226, 0.019701, 0.0) & $[-1.15, 2.2]$ \\
\hline
\end{tabular}
\end{table}

\begin{figure}[h!]
    \centering
    \includegraphics[width=0.6\linewidth]{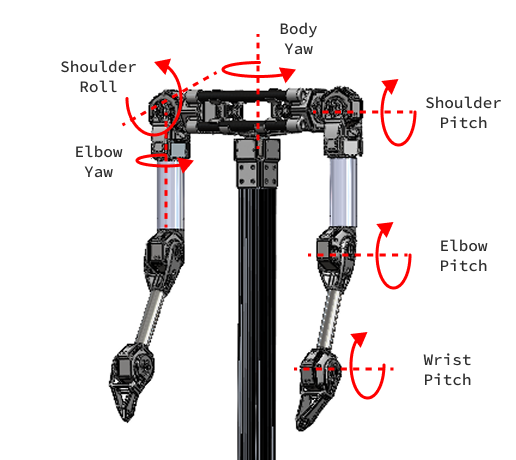}
    \caption{ Coordinate system of Robot B}
    \label{fig:robot_b}
\end{figure}

\section{2D Human Pose Estimation}
\label{appendix:human_pose_estimation}
For 2D human pose estimation, we utilized the OpenMMLab library. In the detection phase, we used the RTMDet-M model (rtmdet\_m\_8xb32-100e\_coco-obj365-person-235e8209.pth). Following detection, we utilized HRNet-W48 (hrnet\_w48\_coco\_wholebody\_384x288\_dark-f5726563\_20200918.pth) for precise keypoint estimation. The HRNet architecture maintains high-resolution representations throughout the network, enabling accurate localization of body joints. This specific model was trained on the COCO-WholeBody dataset with a 384×288 input resolution and incorporates the DARK pose estimation technique for sub-pixel accuracy. The combination of these models provided reliable human pose tracking that was essential for evaluating the physical interaction capabilities of our robotic systems.

\section{Box Pose Estimation}
\label{appendix:box_pose_estimation}
For real-time object detection in our system, we implemented YOLOv8. We created a custom dataset specifically tailored to our application environment, with all annotations performed using the Computer Vision Annotation Tool (CVAT). This approach enabled precise labeling of objects of interest while maintaining consistency across the training dataset. The segmentation model achieves an impressive inference rate of 100 Hz, allowing our system to process visual information at real-time speeds necessary for responsive robotic interaction. This high-frequency detection capability proved essential for tracking dynamic objects in the environment and facilitating accurate decision-making in our experimental scenarios.

\section{Latent Adaptive Planner Algorithms}
\label{appendix:algorithm}
In this section, we provide detailed algorithms for the Latent Adaptive Planner (LAP) framework. We describe both the training procedure and the variational replanning methodology used during inference.
\subsection{Classical Variational Bayes Learning of LAP}

Algorithm \ref{alg:lap_training} outlines the training procedure for our Latent Adaptive Planner using Classical Variational Bayes. This approach optimizes both local parameters (latent plans for individual trajectories) and global parameters (shared decoder model) in an alternating fashion.

\begin{algorithm}[h!]
\caption{LAP: Classical Variational Bayes Learning}
\label{alg:lap_training}
\begin{algorithmic}[1]
\REQUIRE Trajectory dataset $\mathcal{D} = \{\rvx_i\}_{i=1}^N$, where $\rvx_i = \{(o_t^i, a_t^i)\}_{t=1}^T$
\ENSURE Trained model parameters $\theta$
\STATE Initialize global parameters $\theta$ randomly
\REPEAT
    \STATE Sample a mini-batch of trajectories $\{\rvx_i\}_{i=1}^B$ from $\mathcal{D}$
    \FOR{each $\rvx_i$ in the mini-batch}
        \STATE Initialize local parameters $(\boldsymbol{\mu}_i, \boldsymbol{\sigma}_i)$ randomly
        \FOR{$j = 1$ to $T_{\text{local}}$}
            \STATE Sample $\rvz_i \sim \mathcal{N}(\boldsymbol{\mu}_i, \boldsymbol{\sigma}_i^2)$ using reparameterization trick
            \STATE Compute ELBO: $\mathcal{L}(\theta, \boldsymbol{\mu}_i, \boldsymbol{\sigma}_i) = \mathbb{E}_{q(\rvz|\rvx_i)}[\log p_{\theta}(\rvx_i|\rvz)] - \KL(q(\rvz|\rvx_i) \| p(\rvz))$
            \STATE Update $(\boldsymbol{\mu}_i, \boldsymbol{\sigma}_i)$ with gradient ascent on $\mathcal{L}(\theta, \boldsymbol{\mu}_i, \boldsymbol{\sigma}_i)$
        \ENDFOR
    \ENDFOR
    \STATE Sample $\rvz_i \sim \mathcal{N}(\boldsymbol{\mu}_i, \boldsymbol{\sigma}_i^2)$ for each trajectory in mini-batch
    \STATE Update $\theta$ with batch gradient ascent on $\frac{1}{B} \sum_{i=1}^B \mathcal{L}(\theta, \boldsymbol{\mu}_i, \boldsymbol{\sigma}_i)$
\UNTIL convergence
\end{algorithmic}
\end{algorithm}

The key insight of this algorithm is the alternating optimization between local and global parameters. For each trajectory, we first optimize the latent plan distribution parameters $(\boldsymbol{\mu}_i, \boldsymbol{\sigma}_i)$ using $T_{\text{local}}$ steps of gradient ascent (typically 16 steps in our implementation). After optimizing all local parameters in the mini-batch, we then perform a single update of the global parameters $\theta$ based on the average ELBO across the mini-batch.

\subsection{Variational Replanning Algorithm}

Algorithm \ref{alg:variational_replanning} presents our variational replanning approach, which enables efficient adaptation during test time as new observations become available. This approach uses Bayesian updating in the latent space, treating previous distributions as prior beliefs.

\begin{algorithm}[t]
\caption{LAP: Variational Replanning}
\label{alg:variational_replanning}
\begin{algorithmic}[1]
\REQUIRE Trained model parameters $\theta$, replanning horizon $\Delta$, initial observation $o_1$
\ENSURE Robot actions $\{a_t\}_{t=1}^T$
\STATE // Initial planning
\STATE Initialize $\boldsymbol{\mu}_0$ and $\boldsymbol{\sigma}_0$ randomly
\FOR{$j = 1$ to $T_{\text{local}}$}
    \STATE Sample $\rvz \sim \mathcal{N}(\boldsymbol{\mu}_0, \boldsymbol{\sigma}_0^2)$ using reparameterization trick
    \STATE Compute objective: $\mathcal{L}_0 = \log p_{\theta}(\rvx_{0:1}|\rvz) - \KL(\mathcal{N}(\boldsymbol{\mu}_0, \boldsymbol{\sigma}_0^2) \| p(\rvz))$
    \STATE Update $(\boldsymbol{\mu}_0, \boldsymbol{\sigma}_0)$ with gradient ascent on $\mathcal{L}_0$
\ENDFOR
\STATE Sample $\rvz_0 \sim \mathcal{N}(\boldsymbol{\mu}_0, \boldsymbol{\sigma}_0^2)$
\STATE Generate and execute action $a_1 \sim p_{\theta}(a_1|o_1, \rvz_0)$
\STATE $t \leftarrow 1$
\STATE // Adaptive replanning loop
\WHILE{task not complete}
    \STATE Observe $o_{t+1}$
    \IF{$t \bmod \Delta = 0$}
        \STATE // Variational replanning
        \STATE Set $\boldsymbol{\mu}_t \leftarrow \boldsymbol{\mu}_{t-\Delta}$, $\boldsymbol{\sigma}_t \leftarrow \boldsymbol{\sigma}_{t-\Delta}$
        \FOR{$j = 1$ to $T_{\text{replan}}$}
            \STATE Sample $\rvz \sim \mathcal{N}(\boldsymbol{\mu}_t, \boldsymbol{\sigma}_t^2)$ using reparameterization trick
            \STATE Compute objective: \\$\mathcal{L}_t = \log p_{\theta}(\rvx_{t-\Delta+1:t+1}|\rvx_{0:t-\Delta}, \rvz) - \KL(\mathcal{N}(\boldsymbol{\mu}_t, \boldsymbol{\sigma}_t^2) \| \mathcal{N}(\boldsymbol{\mu}_{t-\Delta}, \boldsymbol{\sigma}_{t-\Delta}^2))$
            \STATE Update $(\boldsymbol{\mu}_t, \boldsymbol{\sigma}_t)$ with gradient ascent on $\mathcal{L}_t$
        \ENDFOR
        \STATE Sample $\rvz_t \sim \mathcal{N}(\boldsymbol{\mu}_t, \boldsymbol{\sigma}_t^2)$
    \ENDIF
    \STATE Generate action $a_{t+1} \sim p_{\theta}(a_{t+1}|\rvx_{0:t+1}, \rvz_t)$
    \STATE Execute action $a_{t+1}$
    \STATE $t \leftarrow t + 1$
\ENDWHILE
\end{algorithmic}
\end{algorithm}

The variational replanning algorithm has two key phases. First, we perform initial planning to establish our initial belief about the latent plan. Then, during task execution, we periodically update our belief through Bayesian updating every $\Delta$ timesteps.

A crucial aspect of this approach is that we treat the previously inferred distribution $\mathcal{N}(\boldsymbol{\mu}_{t-\Delta}, \boldsymbol{\sigma}_{t-\Delta}^2)$ as a prior for the current update, as shown in the KL-divergence term of the objective function. This enables efficient adaptation while maintaining temporal consistency in the latent plan.

In practice, we set $T_{\text{replan}} = 1$ for computational efficiency, which is sufficient for incremental updates given that we're starting from a previously optimized distribution. This stands in contrast to the more expensive initial planning phase that uses $T_{\text{local}} = 16$ steps.

\subsection{Implementation Details}

For the variational inference optimization, we use AdamW optimizer with a learning rate of 1e-3 for optimizing local parameters and 2e-4 for global parameters. The latent dimension is set to 64, and we found that diagonal covariance matrices provide a good balance between expressiveness and computational efficiency.

The replanning horizon $\Delta$ is a tunable parameter that trades off computational cost against adaptability. In our experiments, we found $\Delta = 10$ to provide good results, allowing the system to adapt to changing conditions while maintaining real-time performance.

\label{appendix:lap}
\end{document}